\documentclass{article}

\usepackage{times}
\usepackage{graphicx} 
\usepackage{subfigure} 

\usepackage{amsmath}
\usepackage{amsfonts}

\usepackage{natbib}

\usepackage{algorithm}
\usepackage{algorithmic}

\usepackage{hyperref}

\usepackage[accepted]{icml2017}
\usepackage{color}

\usepackage{booktabs}

\icmltitlerunning{Neural Optimizer Search with Reinforcement Learning}

\begin{document} 

\twocolumn[

\icmltitle{Neural Optimizer Search with Reinforcement Learning}



\icmlsetsymbol{equal}{*}

\begin{icmlauthorlist}
\icmlauthor{Irwan Bello}{equal,brain}
\icmlauthor{Barret Zoph}{equal,brain}
\icmlauthor{Vijay Vasudevan}{brain}
\icmlauthor{Quoc V. Le}{brain}
\end{icmlauthorlist}

\icmlaffiliation{brain}{Google Brain}

\icmlcorrespondingauthor{Irwan Bello}{ibello@google.com}
\icmlcorrespondingauthor{Barret Zoph}{barretzoph@google.com}
\icmlcorrespondingauthor{Vijay Vasudevan}{vrv@google.com}
\icmlcorrespondingauthor{Quoc V. Le}{qvl@google.com}

\icmlkeywords{Deep Learning, Reinforcement Learning, Optimization Methods}

\vskip 0.3in
]



\newcommand{\sgn}{\mathrm{sign}}
\newcommand{\clip}{\mathrm{clip}}
\newcommand{\drop}{\mathrm{drop}}
\newcommand{\Adam}{\mathrm{Adam}}
\newcommand{\rmsprop}{\mathrm{RMSProp}}

\printAffiliationsAndNotice{\icmlEqualContribution} 

\begin{abstract} 
We present an approach to automate the process of discovering optimization methods, with a focus on deep learning architectures. We train a Recurrent Neural Network controller to generate a string in a domain specific language that describes a mathematical update equation based on a list of primitive functions, such as the gradient, running average of the gradient, etc.
The controller is trained with Reinforcement Learning to maximize the performance of a model after a few epochs.
On CIFAR-10, our method discovers several update rules that are better than many commonly used optimizers, such as Adam, RMSProp, or SGD with and without Momentum on a ConvNet model. 
We introduce two new optimizers, named PowerSign and AddSign, which we show transfer well and improve training on a variety of different tasks and architectures, including ImageNet classification and Google's neural machine translation system.
\end{abstract} 

\section{Introduction}
\label{sec:intro}
The choice of the right optimization method plays a major role in the success of training deep learning models. Although Stochastic Gradient Descent (SGD) often works well out of the box, more advanced optimization methods such as Adam~\cite{kingma2015adam} or Adagrad~\cite{duchi2011adaptive} can be faster, especially for training very deep networks. Designing optimization methods for deep learning, however, is very challenging due to the non-convex nature of the optimization problems.

\begin{figure}
\includegraphics[width=\columnwidth]{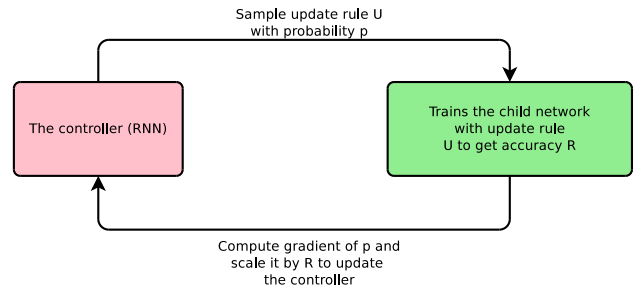}
\caption{An overview of Neural Optimizer Search.}
\label{fig:NOS}
\end{figure}

In this paper, we consider an approach to automate the process of designing update rules for optimization methods, especially for deep learning architectures. The key insight is to use a controller in the form of a recurrent network to generate an update equation for the optimizer. The recurrent network controller is trained with reinforcement learning to maximize the accuracy of a particular model architecture, being trained for a fixed number of epochs with the update rule, on a held-out validation set. This process is shown in Figure~\ref{fig:NOS}.

On CIFAR-10, our approach discovers several update rules that are better than many commonly used optimizers such as Adam, RMSProp, or SGD with and without Momentum on a small ConvNet model. Many of the generated update equations can be easily transferred to new architectures and datasets. For instance, update rules found on a small ConvNet architecture improve training of the Wide ResNet architecture~\cite{zagoruyko2016wide} when compared to Adam, RMSProp, Momentum, and SGD. On ImageNet, our update rules improve top-1 and top-5 accuracy of a state-of-the-art mobile sized model~\cite{zoph2017learning} by up to 0.4\%.
The same update rules also work well on Google's Neural Machine Translation system~\cite{wu2016google}, giving an improvement of up to 0.7 BLEU on the WMT 2014 English to German task.

\section{Related Work}

Neural networks are difficult and slow to train, and many methods have been designed to reduce this difficulty (e.g.,~\citet{riedmiller1992rprop,lecun2012efficient,schraudolph2002fast,martens2010deep,ngiam2011optimization,duchi2011adaptive,zeiler2012adadelta,martens2012training,schaul2013no,pascanu2013revisiting,pascanu2013difficulty,kingma2015adam,ba2017dis}). More recent optimization methods combine insights from both stochastic and batch methods in that they use a small minibatch, similar to SGD, but implement many heuristics to estimate diagonal second-order information, similar to Hessian-free or L-BFGS~\cite{liu1989limited}. This combination often yields faster convergence for practical problems~\cite{duchi2011adaptive,dean2012large,kingma2015adam}. For example, Adam~\cite{kingma2015adam}, a commonly-used optimizer in deep learning, implements simple heuristics to estimate the mean and variance of the gradient, which are used to generate more stable updates during training.

Many of the above update rules are designed by borrowing ideas from convex analysis, even though optimization problems in neural networks are non-convex. Recent empirical results with non-monotonic learning rate heuristics~\cite{loshchilov2016sgdr} suggest that there are still many unknowns in training neural networks and that many ideas in non-convex optimization can be used to improve it.

The goal of our work is to search for better update rules for neural networks in the space of well-known primitives. In other words, instead of hand-designing new update rules from scratch, we use a machine learning algorithm to search among update rules. This goal is shared with recently-proposed methods by~\citet{andrychowicz2016learning,ravi2017opt,wichrowska2017,li2016learning,li2017learning}, which learn to generate numerical updates for training models. The key difference is that our approach generates a mathematical equation for the update instead of numerical updates. The main advantage of generating an equation is that it can easily be transferred to larger tasks and does not require training any additional neural networks for a new optimization problem. Finally, although our method does not aim to optimize the memory usage of update rules, our method discovers update rules that are on par with Adam or RMSProp while requiring less memory.

The concept of using a Recurrent Neural Network for meta-learning has been attempted in the past, either via genetic programming or gradient descent~\cite{schmid92,hochreiter2001learning}. Similar to the above recent methods, these approaches only generate the updates, but not the update equations, as proposed in this paper.

A related approach is using genetic programming to evolve update equations for neural networks (e.g.,~\citet{bengio1994use,evolve_rules, generalized_neural_rule}). Genetic programming however is often slow and requires many heuristics to work well. For that reason, many prior studies in this area have only experimented with very small-scale neural networks. For example, the neural networks used for experiments in~\cite{generalized_neural_rule} have around 100 weights, which is quite small compared to today's standards.

Our approach is reminiscent of recent work in automated model discovery with Reinforcement Learning~\cite{baker2016designing}, especially Neural Architecture Search~\cite{ZophLe}, in which a recurrent network is used to generate the configuration string of neural architectures instead. In addition to applying the key ideas to different applications, this work presents a novel scheme to combine primitive inputs in a much more flexible manner, which makes the search for novel optimizers possible.

Finally, our work is also inspired by the recent studies by~\citet{keskar2016large,zhang2016understanding}, in which it was found that SGD can act as a regularizer that helps generalization. In our work, we use the accuracy on the validation set as the reward signal, thereby implicitly searching for optimizers that can help generalization as well.
\section{Method}

\subsection{A simple domain specific language for update rules}

\begin{figure*}[t!]
\centering
\includegraphics[width=0.7\textwidth]{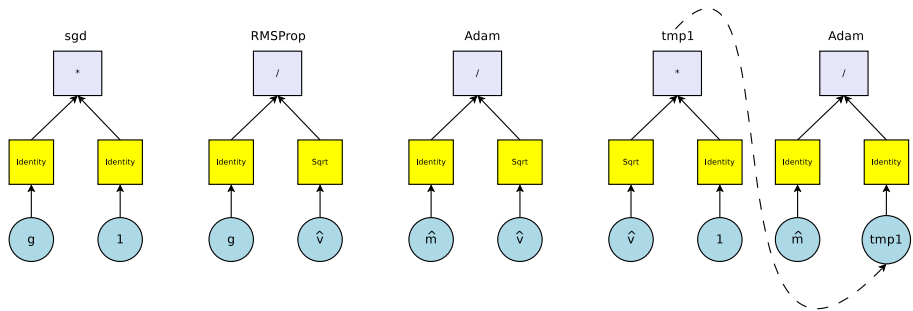}
\caption{Computation graph of some commonly used optimizers: SGD, RMSProp, Adam. Here, we show the computation of Adam in 1 step and 2 steps. Blue boxes correspond to input primitives or temporary outputs, yellow boxes are unary functions and gray boxes represent binary functions. $g$ is the gradient, $\hat{m}$ is the bias-corrected running estimate of the gradient, and $\hat{v}$ is the bias-corrected running estimate of the squared gradient.} 
\label{fig:common_optimizers}
\end{figure*}

\begin{figure*}[tbh]
\centering
\includegraphics[width=0.7\textwidth]{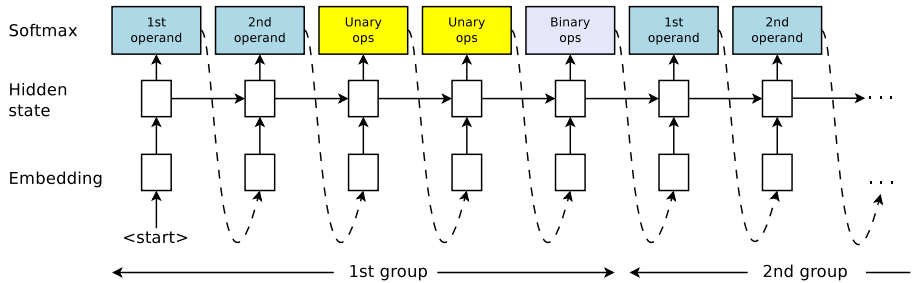}
\caption{Overview of the controller RNN. The controller iteratively selects subsequences of length 5. It first selects the 1st and 2nd operands $op_1$ and $op_2$, then 2 unary functions $u_1$ and $u_2$ to apply to the operands and finally a binary function $b$ that combines the outputs of the unary functions. The resulting $b(u_1(op_1),u_2(op_2))$ then becomes an operand that can be selected in the subsequent group of predictions, or becomes the update rule. Every prediction is carried out by a softmax classifier and then fed into the next time step as input.} 
\label{fig:controller-RNN}
\end{figure*}

In our framework, the controller generates strings corresponding to update rules, which are then applied to a neural network to estimate the update rule's performance. This performance is then used to update the controller so that the controller can generate improved update rules over time.

To map strings sampled by the controller to an update rule, we design a domain specific language that relies on a parenthesis-free notation (in contrast to the classic infix notation).
Our choice of domain specific language (DSL) is motivated by the observation that the computational graph of most common optimizers can be represented as a simple binary expression tree, assuming input primitives such as the gradient or the running average of the gradient and basic unary and binary functions.

We therefore express each update rule with a string describing 1) the first operand to select, 2) the second operand to select, 3) the unary function to apply on the first operand, 4) the unary function to apply on the second operand and 5) the binary function to apply to combine the outputs of the unary functions.
The output of the binary function is then either temporarily stored in our operand bank (so that it can be selected as an operand in subsequent parts of the string) or used as the final weight update as follows:

$$ \Delta w = \lambda * b(u_1(op_1),u_2(op_2))$$

where $op_1$, $op_2$, $u_1(.)$, $u_2(.)$ and $b(.,.)$ are the operands, the unary functions and the binary function corresponding to the string, $w$ is the parameter that we wish to optimize and $\lambda$ is the learning rate.

With a limited number of iterations, our DSL can only represent a subset of all mathematical equations. However we note that it can represent common optimizers within one iteration assuming access to simple primitives. Figure~\ref{fig:common_optimizers} shows how some commonly used optimizers can be represented in the DSL.
We also note that multiple strings in our prediction scheme can map to the same underlying update rule, including strings of different lengths (c.f. the two representations of Adam in Figure~\ref{fig:common_optimizers}). This is both a feature of our action space corresponding to mathematical expressions (addition and multiplication are commutative for example) and our choice of DSL. 

\subsection{Controller optimization with policy gradients}

Our controller is implemented as a Recurrent Neural Network which samples strings of length $5n$ where $n$ is a number of iterations fixed during training (see Figure~\ref{fig:controller-RNN}).
Since the operand bank grows as more iterations are computed, we use different softmax weights at every step of prediction. 

The controller is trained to maximize the performance of its sampled update rules on a specified model. The training objective is formulated as follows:

\begin{equation}
  J(\theta) = \mathbb{E}_{\Delta \sim p_{\theta}(.)} [R(\Delta)]
  \label{eq:rl-obj}
\end{equation}

where $R(\Delta)$ corresponds to the accuracy on a held-out validation set obtained after training a target network with update rule $\Delta$.

\citet{ZophLe} train their controller using a vanilla policy gradient obtained via REINFORCE~\cite{Williams92simplestatistical}, which is known to exhibit poor sample efficiency. We find that using the more sample efficient Proximal Policy Optimization (PPO)~\cite{schulmanppo} algorithm speeds up convergence of the controller. For the baseline function in PPO, we use a simple exponential moving average of previous rewards.

\subsection{Accelerated Training}
\label{sec:param_server}
To further speed up the training of the controller RNN, we employ a distributed training scheme. In our distributed training scheme the samples generated from the controller RNN are added to a queue, and run on a set of distributed workers that are connected across a network. This scheme is different from~\cite{ZophLe} in that now a parameter server and controller replicas are not needed for the controller RNN, which simplifies training. At each iteration, the controller RNN samples a batch of update rules and adds them to the global worker queue. Once the training of the child network is complete, the accuracy on a held-out validation set is computed and returned to the controller RNN, whose parameters get updated with PPO. New samples are then generated and this same process continues.

Ideally, the reward fed to the controller would be the performance obtained when running a model with the sampled optimizer until convergence. However, such a setup requires significant computation and time. To help deal with these issues, we propose the following trade-offs to greatly reduce computational complexity. First, we find that searching for optimizers with a small two layer convolutional network provides enough of a signal for whether an optimizer would do well on much larger models such as the Wide ResNet model. Second, we train each model for a modest 5 epochs only, which also provides enough signal for whether a proposed optimizer is good enough for our needs.
These techniques allow us to run experiments more quickly and efficiently compared to \citet{ZophLe}, with our controller experiments typically converging in less than a day using 100 CPUs, compared to 800 GPUs over several weeks.

\section{Experiments}

\subsection{Search space}

The operands, unary functions and binary functions that are accessible to our controller are the following:
\begin{itemize}
\item {\bf Operands}: $g$, $g^2$, $g^3$, $\hat{m}$, $\hat{v}$, $\hat{\gamma}$, $\sgn(g)$, $\sgn(\hat{m})$, $1$, $2$, $~\epsilon \sim N(0,0.01)$, $10^{-4} w$, $10^{-3} w$, $10^{-2} w$, $10^{-1} w$, Adam and RMSProp.
\item {\bf Unary functions} which map input $x$ to: $x$, $-x$, $e^x$, $\log{|x|}$, $\sqrt{|x|}$, $clip(x, 10^{-5})$, $clip(x, 10^{-4})$, $clip(x, 10^{-3})$, $drop(x, 0.1)$, $drop(x, 0.3)$, $drop(x, 0.5)$ and $\sgn(x)$.
\item {\bf Binary functions} which map $(x,y)$ to $x+y$ (addition), $x-y$ (subtraction), $x*y$ (multiplication), $\frac{x}{y + \delta}$ (division), $x^y$ (exponentiation) or $x$ (keep left).
\end{itemize}

Here, $\hat{m}$, $\hat{v}$, $\hat{\gamma}$ are running exponential moving averages of $g$, $g^2$ and $g^3$, obtained with decay rates $\beta_1$, $\beta_2$ and $\beta_3$ respectively, $drop(.|p)$ sets its inputs to 0 with probability p and $clip(.|l)$ clips its input to $[-l, l]$.
All operations are applied element-wise.

Additionally, we give the controller access to {\bf decay operands} based on the current training step such as:
\begin{itemize}
    \item linear decay: $ld = 1-\frac{t}{T}$.
    \item cyclical decay: $cd_n = 0.5 * (1 + cos(2 \pi n \frac{t}{T}))$.
    \item restart decay: $rd_n = 0.5 * (1 + cos(\pi \frac{(tn) \% T}{T}))$ introduced in~\citet{loshchilov2016sgdr}.
    \item annealed noise: $\epsilon_t \sim N(0,1/(1 + t)^{0.55})$ inspired by~\citet{arvind_noise}.
\end{itemize}

where $t$ is the current training step, $T$ is the total number of training steps and $n$ is a hyperparameter controlling the number of periods in the periodic decays. Note that $cd_{\frac{1}{2}}$ corresponds to cosine decay without restarts~\cite{loshchilov2016sgdr}, which we abbreviate as $cd$.

In our experiments, we use binary trees with depths of 1 to 4 which correspond to strings of length 5, 10, 15 and 20 respectively. The above list of operands, unary functions and binary function is quite large, so to address this issue, we find it helpful to only work with subsets of the operands and functions presented above. This leads to typical search space sizes ranging from $10^{6}$ to $10^{11}$ possible update rules.

We also experiment with several constraints when sampling an update rule, such as forcing the left and right operands to be different at each iteration, and not using addition as the final binary function. An additional constraint added is to force the controller to reuse one of the previously computed operands in the subsequent iterations. The constraints are implemented by manually setting the logits corresponding to the forbidden operands or functions to $-\infty$.

\subsection{Experimental details}

Across all experiments, our controller RNN is trained with the Adam optimizer with a learning rate of $10^{-5}$ and a minibatch size of 5. The controller is a single-layer LSTM with hidden state size $150$ and weights are initialized uniformly at random between -0.08 and 0.08. We also use an entropy penalty to aid in exploration. This entropy penalty coefficient is set to $0.0015$.

The child network architecture that all sampled optimizers are run on is a small two layer 3x3 ConvNet. This ConvNet has 32 filters with ReLU activations and batch normalization applied after each convolutional layer. These child networks are trained on the CIFAR-10 dataset, one of the most benchmarked datasets in deep learning.

The controller is trained on a CPU and the child models are trained using 100 distributed workers which also run on CPUs (see Section~\ref{sec:param_server}). Once a worker receives an optimizer to run from the controller, it performs a basic hyperparameter sweep on the learning rate: $10^{i}$ with i ranging from -5 to 1, with each learning rate being tried for 1 epoch on 10,000 CIFAR-10 training examples. The best learning rate after 1 epoch is then used to train our child network for 5 epochs and the final validation accuracy is reported as a reward to the controller. The child networks have a batch size of 100 and evaluate the update rule on a fixed held-out validation set of 5,000 examples. 
In this setup, training a child model with a sampled optimizer generally takes less than 10 minutes. Experiments typically converge within a day. All experiments are carried out using TensorFlow~\cite{tensorflow}.

The hyperparameter values for the update rules are inspired by standard values used in the literature. We set $\delta$ to $10^{-8}$, $\beta_1$ to 0.9 and $\beta_2 = \beta_3$ to 0.999.

\section{Results}

We now introduce some of the promising update rules discovered by the controller. Note that the mathematical representation of an update rule does not necessarily represent how it was found in our search space (for example $\sgn(g)$ can sampled directly as an operand or can be obtained by applying the sign function to $g$).

\subsection{Discovered optimizers}

\begin{figure}[t!]
\includegraphics[width=\columnwidth]{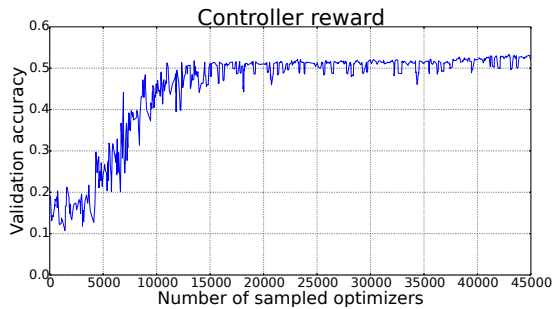}
\caption{Controller reward increasing over time as more optimizers are sampled.} 
\label{fig:controller_reward}
\end{figure}

Our results show that the controller discovers many different updates that perform well when run for 5 epochs on the small ConvNet they were searched over and the maximum accuracy also increases over time. In Figure~\ref{fig:controller_reward}, we show the learning curve of the controller as more optimizers are sampled. 
To filter optimizers that do well when run for many more epochs, we run dozens of our top optimizers for 300 epochs on the Wide ResNet architecture~\cite{zagoruyko2016wide}. To save computational resources we aggressively early stop optimizers that show less promise based on simple heuristics. See Table~\ref{tab:cifar10_appendix} in the Appendix for a few examples of update rules that perform well on the Wide ResNet architecture, but that we did not necessarily experiment with on the other datasets.

Among the regularly sampled formulas is $\sgn(g)*\sgn(m)$, which provides a binary signal on whether the direction of the gradient and its moving average agree on a single dimension. This formula shows up as a sub-component of many of the optimizers sampled by the controller. The controller discovered two fairly intuitive families of update rules based on this formula:

\begin{itemize}
    \item {\bf PowerSign}: $\alpha^{f(t)*\sgn(g)*\sgn(m)}*g$. 
    Some sampled update rules in this family include:
    \begin{itemize}
    \item $e^{\sgn(g)*\sgn(m)}*g$ 
    \item $e^{ld*\sgn(g)*\sgn(m)}*g$
    \item $e^{cd*\sgn(g)*\sgn(m)}*g$
    \item $2^{\sgn(g)*\sgn(m)}*g$
    \end{itemize}
    \item {\bf AddSign}: $(\alpha + f(t)*\sgn(g)*\sgn(m)) * g$.
    Some sampled update rules in this family include:
    \begin{itemize}
    \item $(1 + \sgn(g)*\sgn(m))*g$
    \item $(1 + ld*\sgn(g)*\sgn(m))*g$
    \item $(1 + cd*\sgn(g)*\sgn(m))*g$
    \item $(2 + \sgn(g)*\sgn(m))*g$
    \end{itemize}
\end{itemize}

where $f$ is either $1$ or an internal decay function of the training step $t$ (linear decay, cyclical or restart decays in our experiments).
PowerSign scales the update of each parameter by $\alpha^{f(t)}$ or $1/\alpha^{f(t)}$ depending on whether the gradient and its running moving average agree. AddSign scales the update of each parameter by $\alpha + f(t)$ or $\alpha - f(t)$. For example AddSign without internal decay and $\alpha=1$ only updates parameters for which the gradient and its moving average agree. Note that when $f(t) = 0$, both updates correspond to the usual SGD update, meaning that these internal decays interpolate our update rules to the usual SGD update rule towards the end of training.

Variants of PowerSign and AddSign that replace $g$ with $m$ (or even Adam and RMSProp) are also sampled by the controller but we found these to not perform as well.
Unless specified otherwise, PowerSign specifically refers to $e^{\sgn(g)*\sgn(m)}*g$ (i.e., $\alpha=e$ and no internal decay is applied) and AddSign specifically refers to $(1 + \sgn(g)*\sgn(m)) * g$ (i.e., $\alpha=1$ and no internal decay is applied). When applying internal decay $f$ to the $\sgn(g)*\sgn(m)$ quantity, we refer to our optimizers as PowerSign-f and AddSign-f (e.g., PowerSign-cd and AddSign-ld).
Overall we found our optimizers to be quite robust to slight changes in hyperparameters (including the decay rates for the moving averages). We also found larger values of the base $\alpha$ in the PowerSign optimizer to typically lead to faster convergence with minor final performance loss.

\subsection{Discovered learning rate decays}

The controller also combined our operands into interesting decay functions applied to the learning rate:
\begin{itemize}
    \item $ld * cd$, which we call \emph{linear cosine decay}.
    \item $(ld + \epsilon_t) * cd + 0.001$, which we call \emph{noisy linear decay}.
\end{itemize}

In our experiments, we find that both linear cosine decay and noisy linear cosine decay typically allow for larger initial learning rate and lead to faster convergence than cosine decay~\cite{loshchilov2016sgdr}.

\section{Transferability experiments}

A key advantage of our method of discovering update equations compared to the previous approaches~\cite{andrychowicz2016learning,li2016learning} is that update equations found by our method can be easily transferred to new tasks. In the following experiments, we will use some of the update equations found in the previous experiment on different network architectures and tasks. The controller is not trained again, and the update rules are simply reused. Our goal is to test the transferability of our optimizers on completely different models and tasks.

\subsection{Control Experiment with the Rosenbrock function}
\begin{figure}[h!]
\centering
\includegraphics[width=0.48\columnwidth]{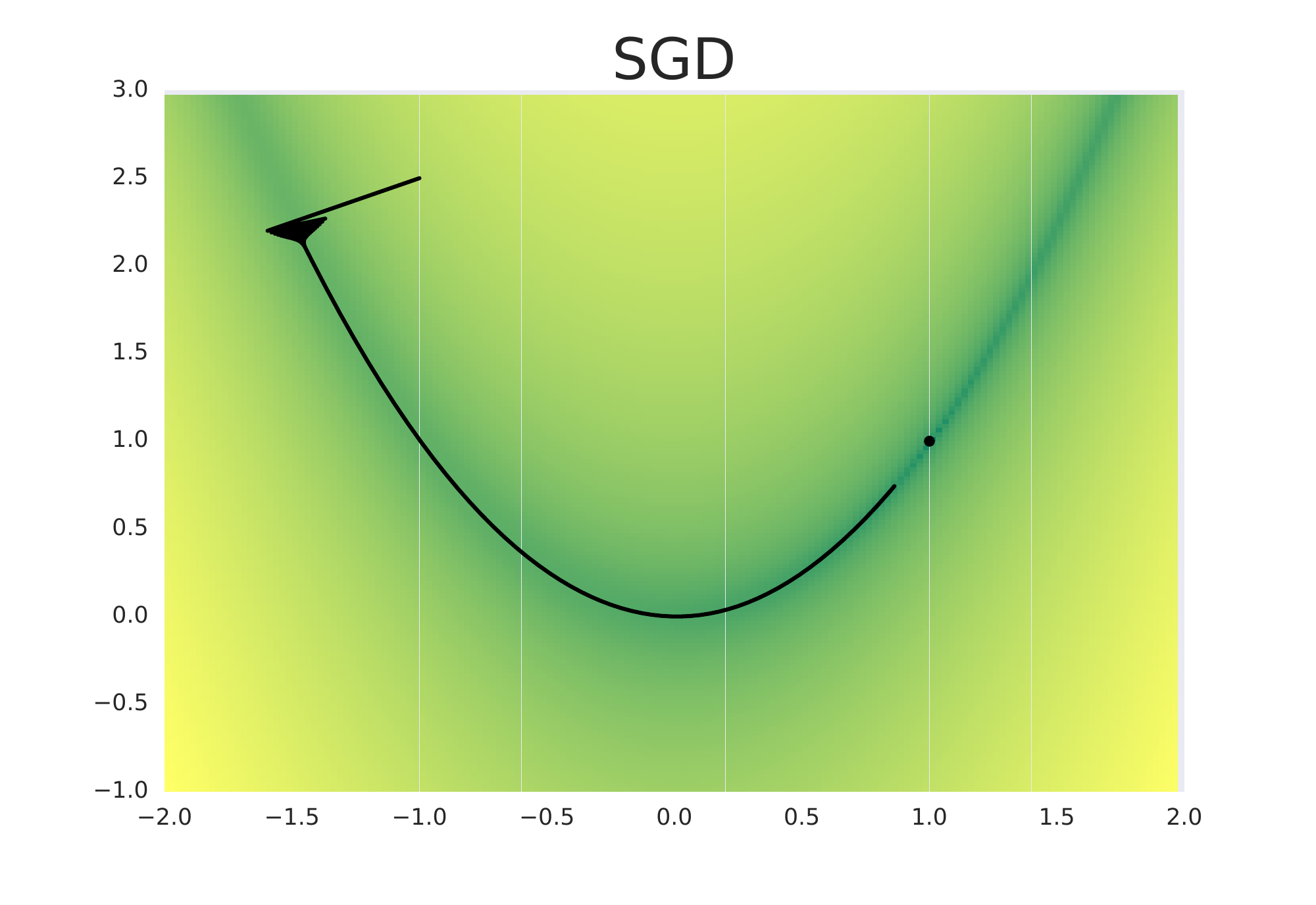} 
\includegraphics[width=0.48\columnwidth]{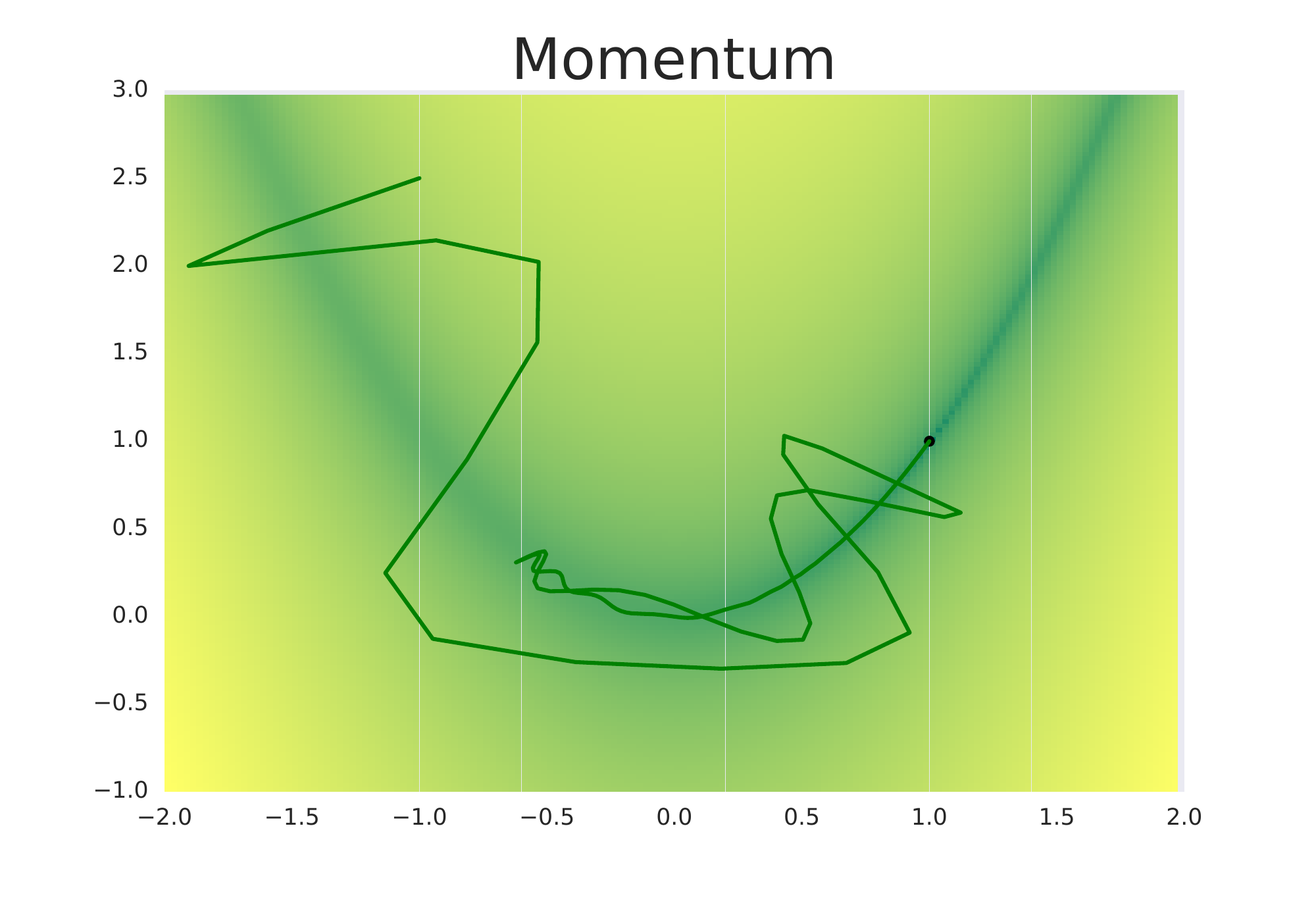}
\includegraphics[width=0.48\columnwidth]{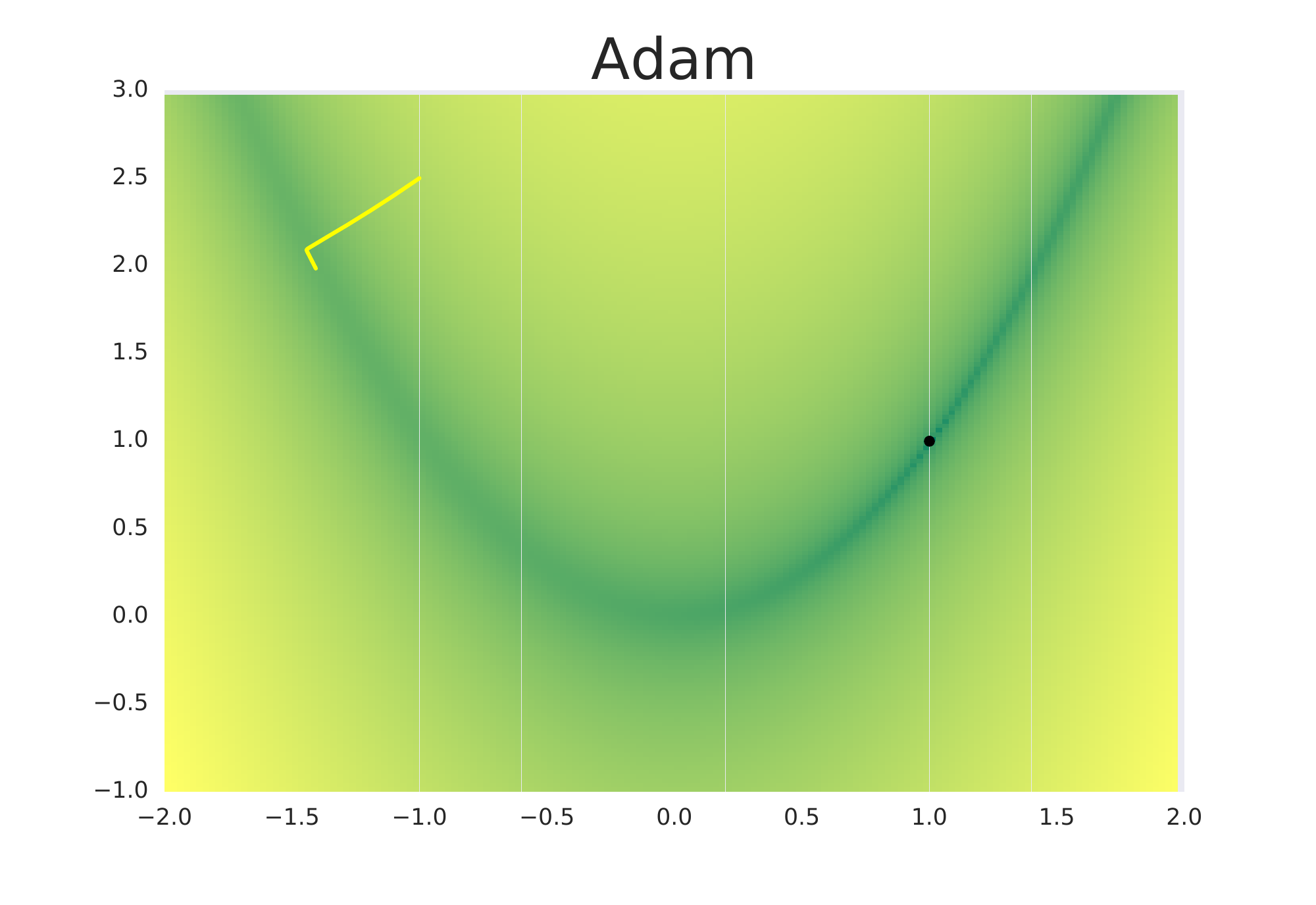}
\includegraphics[width=0.48\columnwidth]{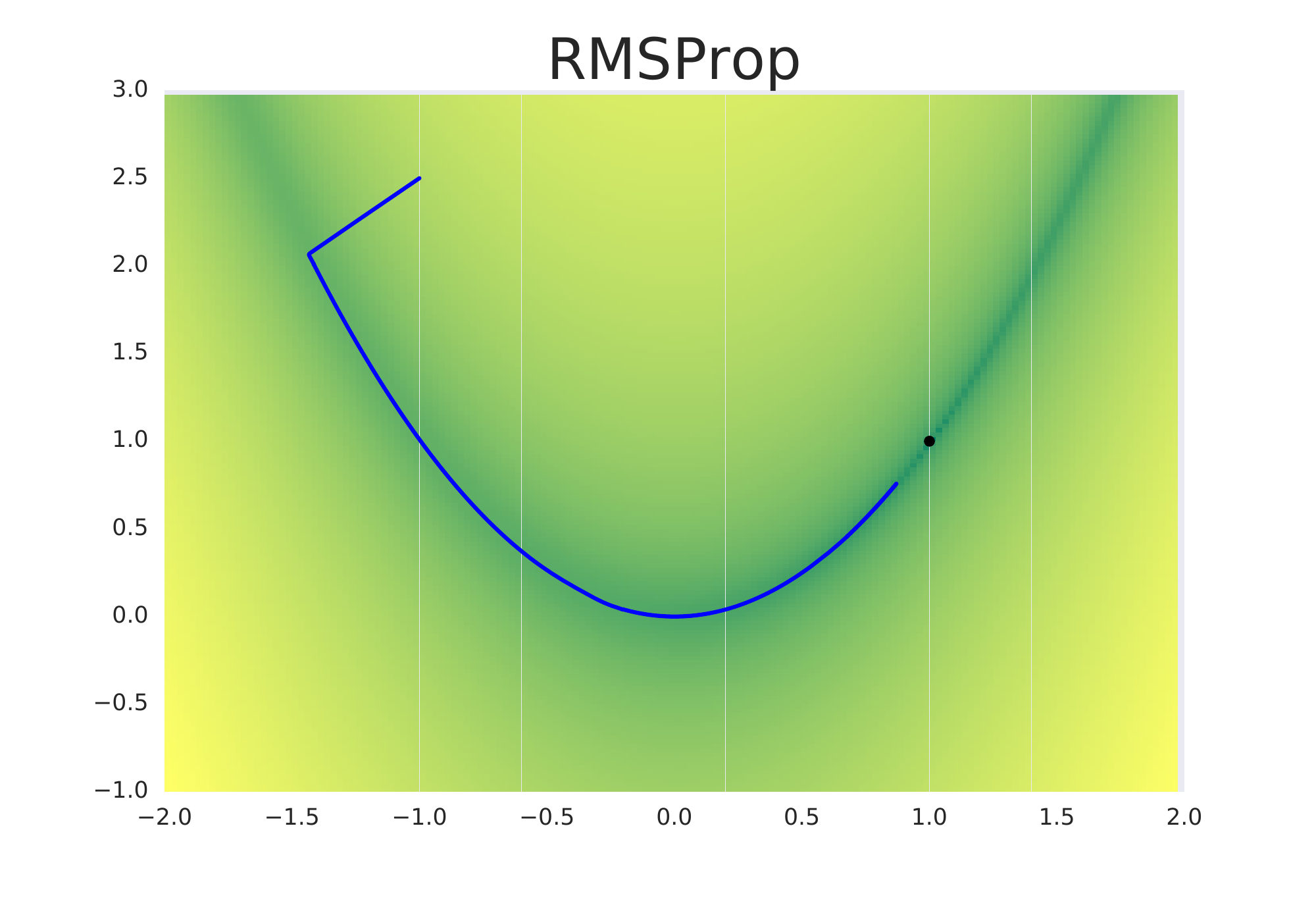} 
\includegraphics[width=0.65\columnwidth]{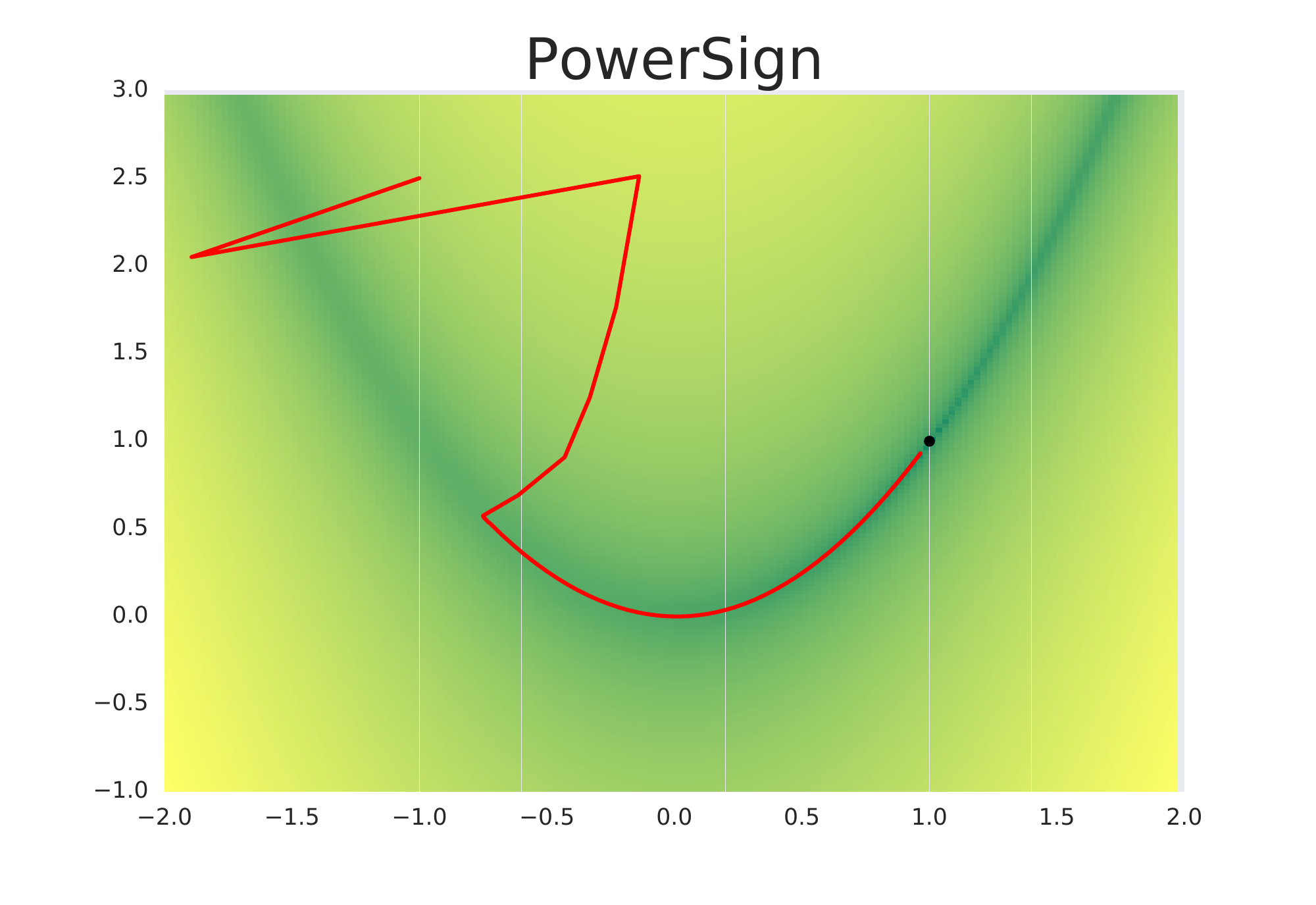} 
\caption{Comparison of the PowerSign optimizer against SGD, Momentum, RMSProp and Adam on the Rosenbrock function. The black dot is the optimum.}
\label{fig:Rosenbrock}
\end{figure}

We first test the PowerSign optimizer on the well-known Rosenbrock function, a common testbed in stochastic optimization and compare its performance against the commonly used deep learning optimizers in TensorFlow~\cite{tensorflow}: Adam, SGD, RMSProp and Momentum. In this experiment, each optimizer is run for 4000 iterations with 4 different learning rates searched over a logarithmic scale and the best performance is plotted (the value of $\epsilon$ for the division in Adam was additionally tuned on a logarithmic scale between $10^{-3}$ and $10^{-8}$). The results in Figure~\ref{fig:Rosenbrock} show that our optimizer outperforms Adam, RMSProp, SGD, and is close to matching the performance of Momentum on this task.

\subsection{CIFAR-10 with the Wide ResNet architecture}

We further investigate the generalizability of the found update rules on a different and much larger model: a Wide ResNet architecture~\cite{zagoruyko2016wide} with 9 million parameters. 
For each commonly used optimizer, we tuned the weight decay and tried 7 different learning rates on a logarithmic scale. For learning rate decays we considered stepwise and cosine decay~\cite{loshchilov2016sgdr}. We found the latter to consistently perform better. For our optimizers we do not use weight decay and try 3 learning rates centered around the one that yielded the best results during the controller search.

\begin{figure}[h!]
\centering
\includegraphics[width=\columnwidth]{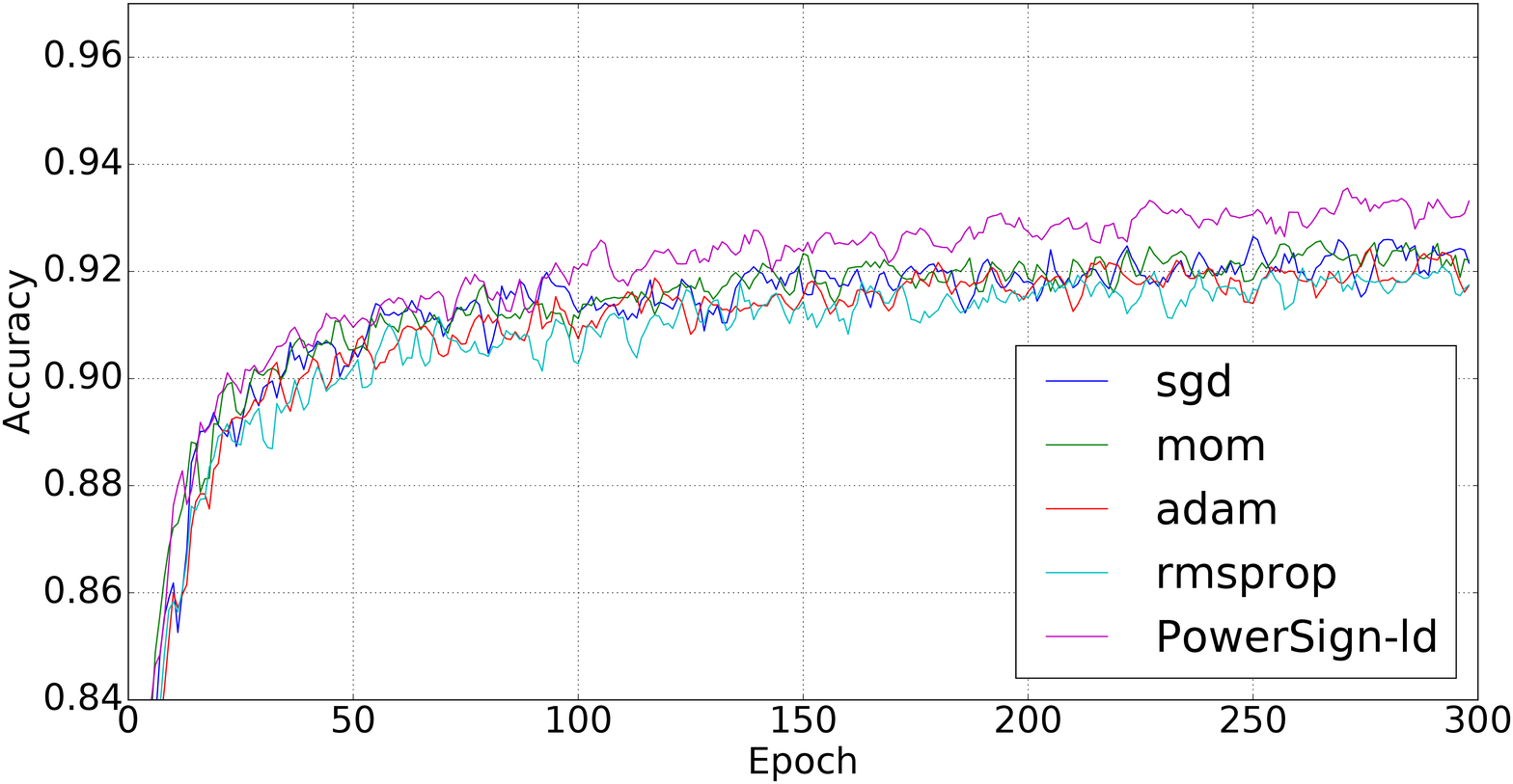} \\
\includegraphics[width=\columnwidth]{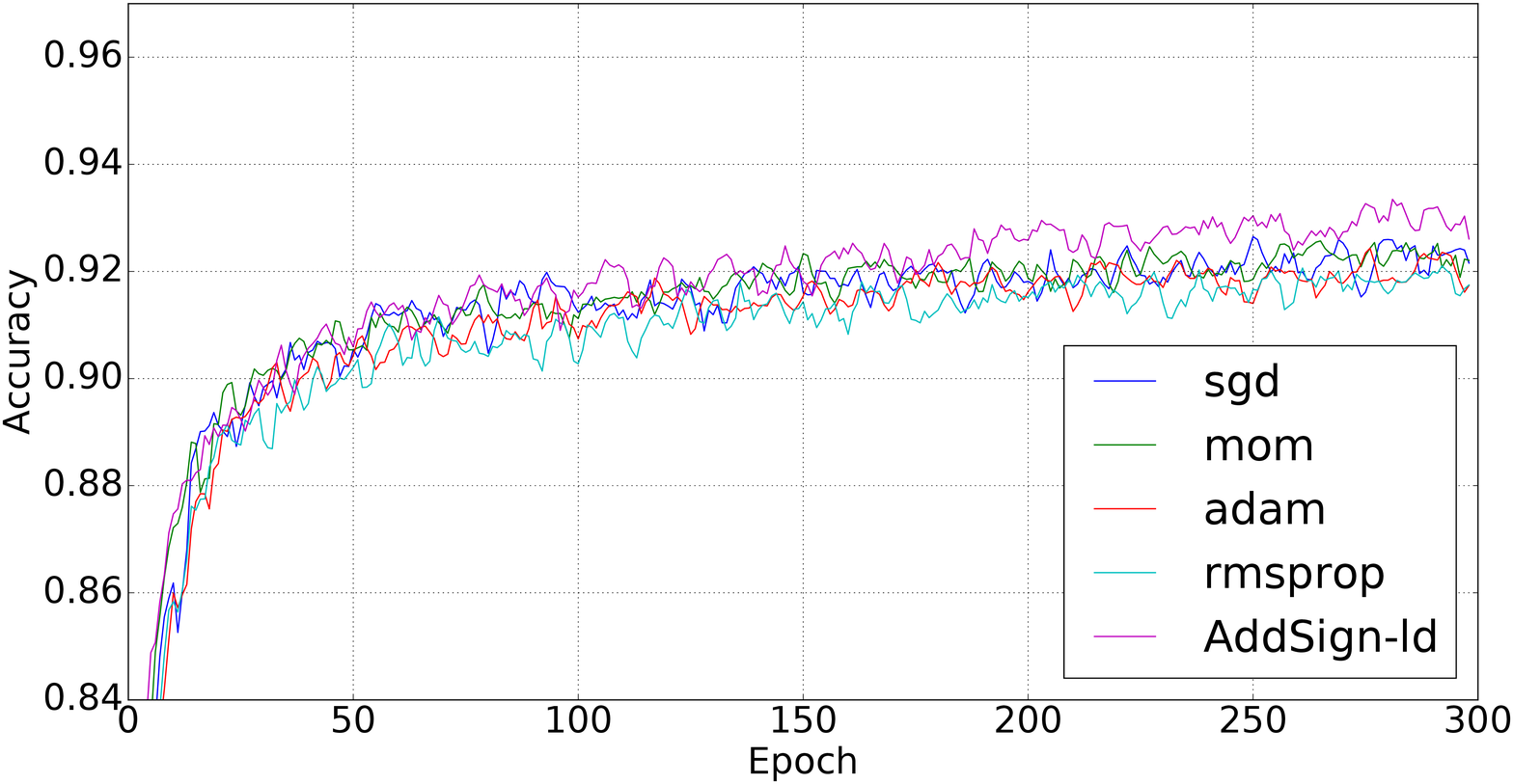} \\
\caption{Comparison of the PowerSign-ld and AddSign-ld optimizers, found with Neural Optimizer Search, on the Wide ResNet architecture. No learning rate decay is applied. Best viewed in color.}
\label{fig:best_opt_plot_wrn_nd}
\end{figure}

When not applying learning rate decay, some of our optimizers outperform common optimizers by a size-able margin (see Table~\ref{tab:cifar10_nodecay} and Figure~\ref{fig:best_opt_plot_wrn_nd}), especially when employing internal decays. This feature of our optimizers can be advantageous in tasks that do not necessarily require learning rate decays.

\begin{table}[h!]
\centering
\small
\begin{tabular}{l|cc}
\toprule
\multicolumn{1}{c|}{\bf Optimizer} & {\bf Best Test} & {\bf Final Test}\\ 
\midrule
SGD & 93.0 & 92.3  \\
Momentum & 93.0 & 92.2 \\
Adam & 92.6 & 92.3 \\
RMSProp & 92.3 & 91.6 \\
\midrule
PowerSign & 93.0 & 92.4 \\
PowerSign-ld & 93.6 & 93.4 \\
PowerSign-cd & 93.7 & 93.1 \\
PowerSign-rd$_{10}$ & 94.2 & 92.6 \\
PowerSign-rd$_{20}$ & 94.4 & 92.0 \\
AddSign & 93.0 & 92.6 \\
AddSign-ld & 93.5 & 92.0 \\
AddSign-cd & 93.6 & 92.4 \\
AddSign-rd$_{10}$ & 94.2 & 94.0 \\
{\bf AddSign-rd$_{20}$} & {\bf 94.4} & {\bf 94.3} \\
\bottomrule
\end{tabular}%
\caption{Performance of the PowerSign and AddSign optimizers against standard optimizers on the Wide-ResNet architecture on CIFAR-10 without learning rate decay. Results are averaged over 5 runs.}
\label{tab:cifar10_nodecay}
\end{table}

\begin{figure}[h!]
\centering
\includegraphics[width=\columnwidth]{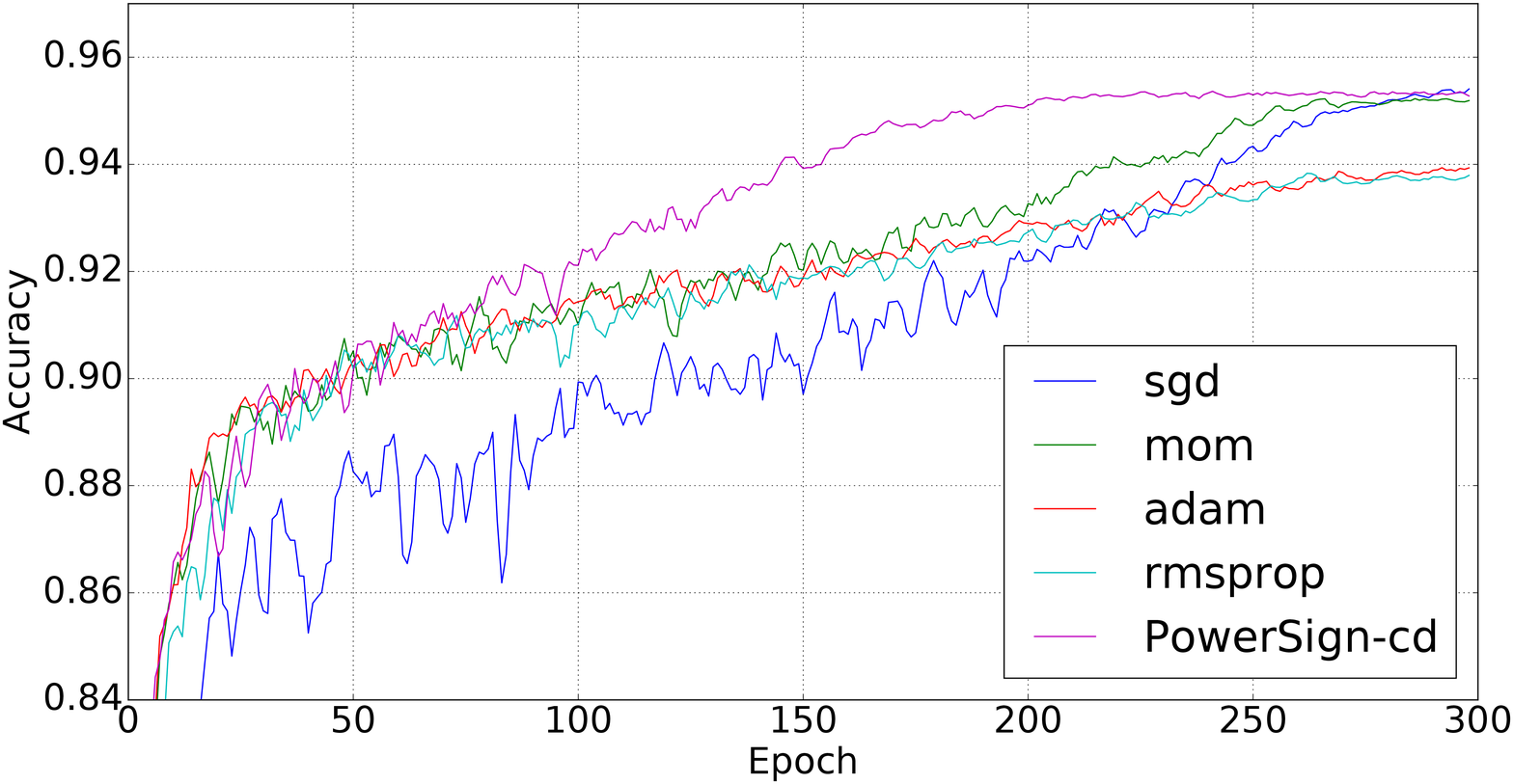} \\
\includegraphics[width=\columnwidth]{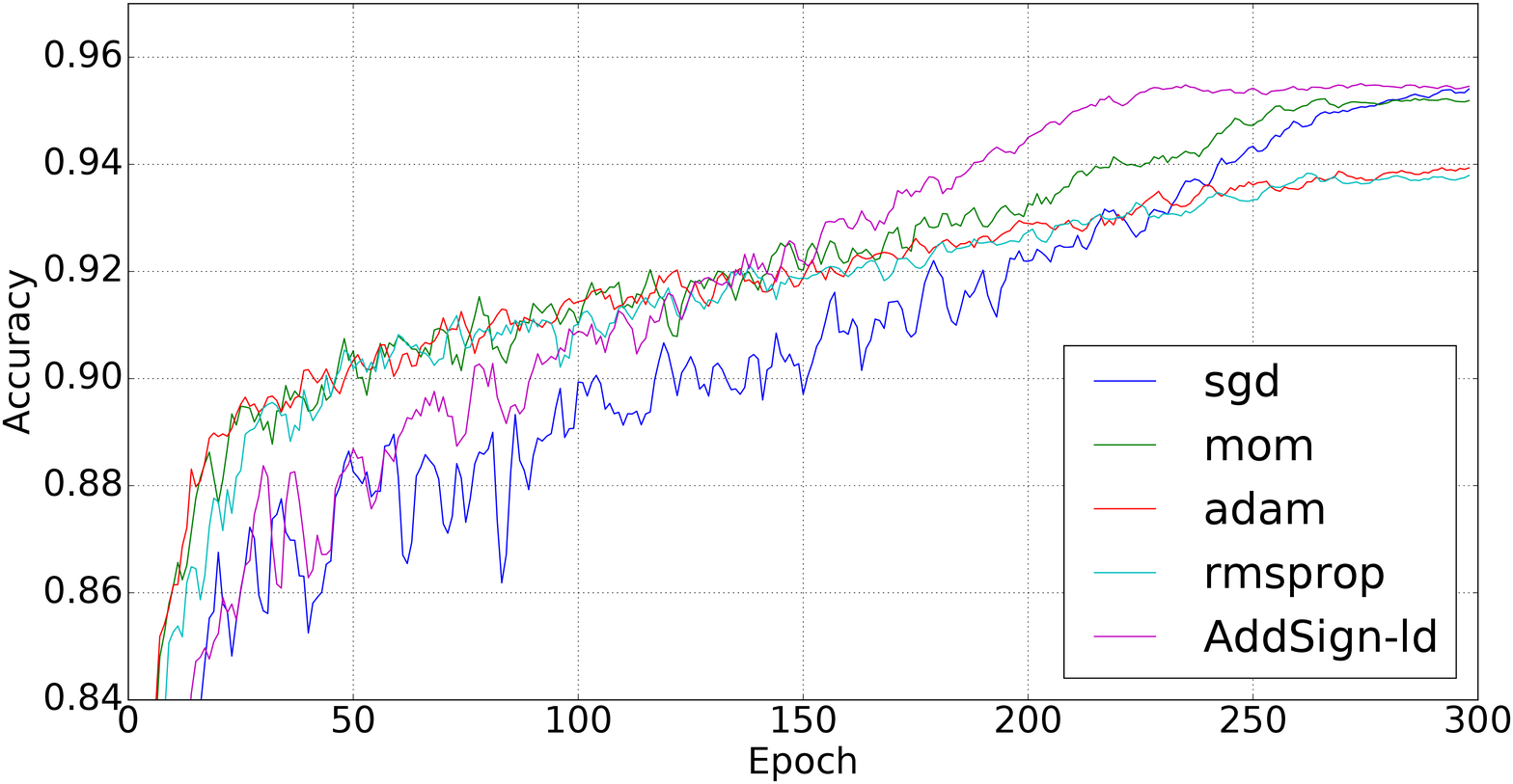} \\
\caption{Comparison of the PowerSign-cd and AddSign-ld optimizers, found with Neural Optimizer Search, on the Wide ResNet architecture. The PowerSign and AddSign optimizers are applied with linear cosine decay. SGD, Momentum, RMSProp and Adam are applied with cosine learning rate decay. Best viewed in color.}
\label{fig:best_opt_plot_wrn}
\end{figure}

Table~\ref{tab:cifar10_decay} shows the comparison between SGD, Adam, RMSProp, Momentum and our selected optimizers when allowing for learning rate decay. 
On this task, our optimizers outperform Adam, RMSProp and Momentum and are on par with SGD. Note that the variance in our optimizers' results may be explained by how good of a learning rate we chose, rather than the superiority of one optimizer over others. We observed that our optimizers tend to converge faster than Momentum/SGD. Interestingly, we also found that the linear cosine decay generally allows for a larger initial learning rate and leads to faster convergence (see Figure~\ref{fig:best_opt_plot_wrn}).

\begin{table}[h!]
\centering
\small
\begin{tabular}{l|cc}
\toprule
\multicolumn{1}{c|}{\bf Optimizer} & {\bf Best Test} & {\bf Final Test}\\ 
\midrule
SGD (cosine) & 95.4 & 95.3  \\
Momentum (cosine) & 95.2 & 95.1 \\
Adam (cosine) & 93.9 & 93.8 \\
RMSProp (cosine) & 93.6 & 93.5 \\
\midrule
AddSign-ld (cosine) & 95.4 & 95.3 \\
AddSign-ld (linear cosine) & 95.3 & 95.1 \\
{\bf AddSign-ld (noisy linear cosine)} & {\bf 95.5} & {\bf 95.4} \\
AddSign-cd (cosine) & 95.3 & 95.1 \\
AddSign-cd (linear cosine) & 95.3 & 95.2 \\
AddSign-cd (noisy linear cosine) & 95.3 & 95.2 \\
PowerSign (cosine) & 95.4 & 95.3 \\
PowerSign (linear cosine) & 95.4 & 95.2 \\
PowerSign-ld (cosine) & 95.3 & 95.2 \\
PowerSign-ld (linear cosine) & 95.3 & 95.2 \\
PowerSign-cd (cosine) & 95.4 & 95.2 \\
PowerSign-cd (linear cosine) & 95.4 & 95.2 \\
\bottomrule
\end{tabular}
\caption{Performance of the PowerSign and AddSign optimizers against standard optimizers on the Wide-ResNet architecture on CIFAR-10 with learning rate decay (specified in parenthesis). Results are averaged over 5 runs.}
\label{tab:cifar10_decay}
\end{table}

\subsection{ImageNet Classification}

We additionally evaluate the PowerSign and AddSign optimizers on the ImageNet classification task. The model we use is the state-of-the-art Mobile sized NASNet-A model from \citet{zoph2017learning}.
We observe an improvement in using the optimizers over the default RMSProp optimizer, in spite of the training pipeline being initially tuned for RMSProp.

\begin{table}[h!]
\centering
\small
\begin{tabular}{l|cc}
\toprule
\multicolumn{1}{l|}{\bf Optimizer} & {\bf Top-1 Accuracy} & {\bf Top-5 Accuracy} \\
\midrule
RMSProp & 73.5 & 91.5 \\
\midrule
{\bf PowerSign-cd} & {\bf 73.9} & {\bf 91.9} \\
AddSign-ld & 73.8 & 91.6 \\
\bottomrule
\end{tabular}
\caption{Performance of our PowerSign and AddSign optimizers against RMSProp on a state-of-the art MobileNet baseline~\cite{zoph2017learning}. All optimizers are applied with cosine learning rate decay.}
\label{tab:mobile_net}
\end{table}
For our optimizers we only tried three different learning rates: 1.0, 2.5, 4.0. All models, including the RMSProp baseline, were trained with cosine learning rate annealing for a fixed 250K training steps with 50 synchronous replicas, each having a batch size of 32. See~\citet{zoph2017learning} for the rest of the training details. We can see from the results in Table~\ref{tab:mobile_net} that both the PowerSign and AddSign achieve improvements over the RMSProp optimizer, with the PowerSign doing the best achieving a 0.4\% top1 and 0.4\% top5 accuracy improvement along with a speedup in convergence.

\subsection{Neural Machine Translation}

We evaluate the PowerSign optimizer on the WMT 2014 English $\rightarrow$ German task. Our optimizer in this experiment is compared against the Adam optimizer~\cite{kingma2015adam}. The architecture of interest is the Google Neural Machine Translation (GNMT) model~\cite{wu2016google}, which was shown to achieve competitive translation quality on the English $\rightarrow$ German task. The GNMT network comprises 8 LSTM layers for both its encoder and decoder~\cite{hochreiter1997long}, with the first layer of the encoder having bidirectional connections. This GNMT model also employs attention in the form of a 1 layer neural network. 
The model is trained in a distributed fashion using a parameter server. Twelve workers are used, with each worker using 8 GPUs and a minibatch size of 128. Further details for this model can be found in \citet{wu2016google}. 

The only change we make to training is to replace Adam with the PowerSign update rule. We note that the GNMT model's hyperparameters, such as weight initialization, were previously tuned to work well with Adam~\cite{wu2016google}, so we expect more tuning can further improve the results of our new update rule.

The results in Table~\ref{tab:gnmt} show that our optimizer does indeed generalize well and achieves an improvement of 0.1 perplexity, which is considered to be a decent gain on this task. This gain in training perplexity enables the model to obtain a 0.5 BLEU improvement over the Adam optimizer on the test set~\citet{wu2016google}. On the validation set, the averaged improvement of 5 points near the peak values is 0.7 BLEU.

\begin{table}[h!]
\centering
\small
\begin{tabular}{l|cc}
\toprule
\multicolumn{1}{c|}{\bf Optimizer} & {\bf Train perplexity} & {\bf Test BLEU} \\ 
\midrule
Adam & 1.49 & 24.5 \\
\midrule
{\bf PowerSign} & {\bf 1.39} & {\bf 25.0} \\
\bottomrule
\end{tabular}%
\caption{Performance of our optimizer against Adam on a strong baseline GNMT model on WMT 2014 English $\rightarrow$ German.}
\label{tab:gnmt}
\end{table}

Finally, the PowerSign update rule is also more memory efficient as it only keeps one running average per parameter, compared to two running averages for Adam. This has practical implications for much larger translation models where Adam cannot currently be used due to memory constraints \cite{largenetworks}.

\subsection{Language Modeling}

Our last experiment is to evaluate our optimizers on the Penn Treebank (PTB) language modeling task~\cite{ptb_marcus,ptb_mikolov}.
Our base model is a 10M parameters single layer LSTM with variational dropout~\cite{variational_gal} and weight tying~\cite{weight_inan,weight_press}. Weight decay, embedding, state and output dropout, state reset probability and embedding ratio are tuned with a black-box hyperparameter tuner~\cite{vizier}, similarly to~\citet{language_model_dm}. We clip the gradient norm to 5.0 and train our model for 150000 steps with a batch size of 64 and a truncated backpropagation length of 35 timesteps.
We compare the PowerSign and AddSign against Adam and SGD in Table~\ref{tab:ptb}. In this setup, our optimizers perform much better than SGD but are outperformed by Adam. Hyperparameter ranges were first manually tuned by experimenting with Adam, so it is possible that this could negatively affect performance for SGD and our optimizers.

\begin{table}[h!]
\centering
\small
\begin{tabular}{l|cc}
\toprule
\multicolumn{1}{l|}{\bf Optimizer} & {\bf Valid perplexity} & {\bf Test Perplexity} \\
\midrule
{\bf Adam} & {\bf 65.7} & {\bf 63.3} \\
SGD & 71.7 & 68.3 \\
\midrule
PowerSign & 68.5 & 64.8 \\
AddSign & 68.8 & 65.2 \\
\bottomrule
\end{tabular}
\caption{Performance of our optimizers against Adam and SGD on the PTB language modeling task. All optimizers are applied with cosine learning rate decay.}
\label{tab:ptb}
\end{table}

\section{Conclusion}

This paper considers an approach for automating the discovery of optimizers with a focus on deep neural network architectures.
One strength of our approach is that it naturally encompasses the environment in which the optimization process happens. One may for example use our method for discovering optimizers that perform well in scenarios where computations are only carried out using 4 bits, or a distributed setup where workers can only communicate a few bits of information to a shared parameter server. 

Unlike previous approaches in learning to learn, the update rules in the form of equations can be easily transferred to other optimization tasks.

Our method discovered two new intuitive update rules, PowerSign and AddSign, that obtain competitive performance against common optimizers on a variety of tasks and models, from image classification with ConvNets to machine translation with LSTMs. Our method also identified a new learning rate annealing scheme, \emph{linear cosine decay}, which we found generally leads to faster convergence than cosine annealing.

In addition to opening up new ways to design update rules, our update rules can now be used to improve the training of deep networks.

\section*{Acknowledgements}
We thank Samy Bengio, Jeff Dean, Vishy Tirumalashetty, David Dohan and the Google Brain team for the help with the project.

\bibliography{main}

\begin{thebibliography}{46}
\providecommand{\natexlab}[1]{#1}
\providecommand{\url}[1]{\texttt{#1}}
\expandafter\ifx\csname urlstyle\endcsname\relax
  \providecommand{\doi}[1]{doi: #1}\else
  \providecommand{\doi}{doi: \begingroup \urlstyle{rm}\Url}\fi

\bibitem[Abadi et~al.(2016)Abadi, Barham, Chen, Chen, Davis, Dean, Devin,
  Ghemawat, Irving, Isard, Kudlur, Levenberg, Monga, Moore, Murray, Steiner,
  Tucker, Vasudevan, Warden, Wicke, Yu, , and Zheng]{tensorflow}
Abadi, Mart{\'\i}n, Barham, Paul, Chen, Jianmin, Chen, Zhifeng, Davis, Andy,
  Dean, Jeffrey, Devin, Matthieu, Ghemawat, Sanjay, Irving, Geoffrey, Isard,
  Michael, Kudlur, Manjunath, Levenberg, Josh, Monga, Rajat, Moore, Sherry,
  Murray, Derek~G., Steiner, Benoit, Tucker, Paul, Vasudevan, Vijay, Warden,
  Pete, Wicke, Martin, Yu, Yuan, , and Zheng, Xiaoqiang.
\newblock Tensorflow: A system for large-scale machine learning.
\newblock \emph{Proceedings of the 12th USENIX Symposium on Operating Systems
  Design and Implementation (OSDI)}, 2016.

\bibitem[Andrychowicz et~al.(2016)Andrychowicz, Denil, Gomez, Hoffman, Pfau,
  Schaul, and de~Freitas]{andrychowicz2016learning}
Andrychowicz, Marcin, Denil, Misha, Gomez, Sergio, Hoffman, Matthew~W., Pfau,
  David, Schaul, Tom, and de~Freitas, Nando.
\newblock Learning to learn by gradient descent by gradient descent.
\newblock In \emph{Advances in Neural Information Processing Systems}, pp.\
  3981--3989, 2016.

\bibitem[Ba et~al.(2017)Ba, Grosse, and Martens]{ba2017dis}
Ba, Jimmy, Grosse, Roger, and Martens, James.
\newblock Distributed second-order optimization using {K}ronecker-factored
  approximations.
\newblock In \emph{International Conference on Learning Representations}, 2017.

\bibitem[Baker et~al.(2016)Baker, Gupta, Naik, and Raskar]{baker2016designing}
Baker, Bowen, Gupta, Otkrist, Naik, Nikhil, and Raskar, Ramesh.
\newblock Designing neural network architectures using reinforcement learning.
\newblock \emph{arXiv preprint arXiv:1611.02167}, 2016.

\bibitem[Bengio et~al.(1994)Bengio, Bengio, and Cloutier]{bengio1994use}
Bengio, Samy, Bengio, Yoshua, and Cloutier, Jocelyn.
\newblock Use of genetic programming for the search of a new learning rule for
  neural networks.
\newblock In \emph{Evolutionary Computation, 1994. IEEE World Congress on
  Computational Intelligence., Proceedings of the First IEEE Conference on},
  pp.\  324--327. IEEE, 1994.

\bibitem[Dean et~al.(2012)Dean, Corrado, Monga, Chen, Devin, Mao, Senior,
  Tucker, Yang, Le, et~al.]{dean2012large}
Dean, Jeffrey, Corrado, Greg, Monga, Rajat, Chen, Kai, Devin, Matthieu, Mao,
  Mark, Senior, Andrew, Tucker, Paul, Yang, Ke, Le, Quoc~V, et~al.
\newblock Large scale distributed deep networks.
\newblock In \emph{Advances in Neural Information Processing Systems}, pp.\
  1223--1231, 2012.

\bibitem[Duchi et~al.(2011)Duchi, Hazan, and Singer]{duchi2011adaptive}
Duchi, John, Hazan, Elad, and Singer, Yoram.
\newblock Adaptive subgradient methods for online learning and stochastic
  optimization.
\newblock \emph{Journal of Machine Learning Research}, 2011.

\bibitem[Gal \& Ghahramani(2016)Gal and Ghahramani]{variational_gal}
Gal, Yarin and Ghahramani, Zoubin.
\newblock A theoretically grounded application of dropout in recurrent neural
  networks.
\newblock In \emph{Advances in Neural Information Processing Systems}, 2016.

\bibitem[Golovin et~al.(2017)Golovin, Solnik, Moitra, Kochanski, Karro, and
  Sculley]{vizier}
Golovin, Daniel, Solnik, Benjamin, Moitra, Subhodeep, Kochanski, Greg, Karro,
  John~Elliot, and Sculley, D.
\newblock Google vizier: A service for black-box optimization.
\newblock In \emph{Proceedings of the 23rd ACM SIGKDD International Conference
  on Knowledge Discovery and Data Mining}, 2017.

\bibitem[Hochreiter \& Schmidhuber(1997)Hochreiter and
  Schmidhuber]{hochreiter1997long}
Hochreiter, Sepp and Schmidhuber, J{\"u}rgen.
\newblock Long short-term memory.
\newblock \emph{Neural Computation}, 9\penalty0 (8):\penalty0 1735--1780, 1997.

\bibitem[Hochreiter et~al.(2001)Hochreiter, Younger, and
  Conwell]{hochreiter2001learning}
Hochreiter, Sepp, Younger, A~Steven, and Conwell, Peter~R.
\newblock Learning to learn using gradient descent.
\newblock In \emph{International Conference on Artificial Neural Networks},
  pp.\  87--94. Springer, 2001.

\bibitem[Inan et~al.(2016)Inan, Khosravi, and Socher]{weight_inan}
Inan, Hakan, Khosravi, Khashayar, and Socher, Richard.
\newblock Tying word vectors and word classifiers: {A} loss framework for
  language modeling.
\newblock In \emph{International Conference on Learning Representations}, 2016.

\bibitem[Keskar et~al.(2016)Keskar, Mudigere, Nocedal, Smelyanskiy, and
  Tang]{keskar2016large}
Keskar, Nitish~Shirish, Mudigere, Dheevatsa, Nocedal, Jorge, Smelyanskiy,
  Mikhail, and Tang, Ping Tak~Peter.
\newblock On large-batch training for deep learning: Generalization gap and
  sharp minima.
\newblock In \emph{International Conference on Learning Representations}, 2016.

\bibitem[Kingma \& Ba(2015)Kingma and Ba]{kingma2015adam}
Kingma, Diederik~P. and Ba, Jimmy.
\newblock Adam: {A} method for stochastic optimization.
\newblock In \emph{International Conference on Learning Representations}, 2015.

\bibitem[Le et~al.(2011)Le, Ngiam, Coates, Lahiri, Prochnow, and
  Ng]{ngiam2011optimization}
Le, Quoc~V., Ngiam, Jiquan, Coates, Adam, Lahiri, Ahbik, Prochnow, Bobby, and
  Ng, Andrew~Y.
\newblock On optimization methods for deep learning.
\newblock In \emph{Proceedings of the 28th International Conference on Machine
  Learning}, 2011.

\bibitem[LeCun et~al.(1998)LeCun, Bottou, Orr, and
  M{\"u}ller]{lecun2012efficient}
LeCun, Yann~A, Bottou, L{\'e}on, Orr, Genevieve~B., and M{\"u}ller,
  Klaus-Robert.
\newblock Efficient backprop.
\newblock In \emph{Neural networks: Tricks of the trade}. Springer, 1998.

\bibitem[Li \& Malik(2016)Li and Malik]{li2016learning}
Li, Ke and Malik, Jitendra.
\newblock Learning to optimize.
\newblock In \emph{International Conference on Learning Representations}, 2016.

\bibitem[Li \& Malik(2017)Li and Malik]{li2017learning}
Li, Ke and Malik, Jitendra.
\newblock Learning to optimize neural nets.
\newblock \emph{arXiv preprint arXiv:1703.00441}, 2017.

\bibitem[Liu \& Nocedal(1989)Liu and Nocedal]{liu1989limited}
Liu, Dong~C and Nocedal, Jorge.
\newblock On the limited memory {BFGS} method for large scale optimization.
\newblock \emph{Mathematical programming}, 45\penalty0 (1):\penalty0 503--528,
  1989.

\bibitem[Loshchilov \& Hutter(2017)Loshchilov and Hutter]{loshchilov2016sgdr}
Loshchilov, Ilya and Hutter, Frank.
\newblock {SGDR}: stochastic gradient descent with restarts.
\newblock In \emph{International Conference on Learning Representations}, 2017.

\bibitem[Marcus et~al.(1993)Marcus, Marcinkiewicz, and Santorini]{ptb_marcus}
Marcus, Mitchell~P, Marcinkiewicz, Mary~Ann, and Santorini, Beatrice.
\newblock Building a large annotated corpus of english: The penn treebank.
\newblock In \emph{Computational Linguistics 19(2):313–330}, 1993.

\bibitem[Martens(2010)]{martens2010deep}
Martens, James.
\newblock Deep learning via {H}essian-free optimization.
\newblock In \emph{Proceedings of the 27th International Conference on Machine
  Learning}, pp.\  735--742, 2010.

\bibitem[Martens \& Sutskever(2012)Martens and Sutskever]{martens2012training}
Martens, James and Sutskever, Ilya.
\newblock Training deep and recurrent networks with {H}essian-free
  optimization.
\newblock In \emph{Neural networks: Tricks of the trade}, pp.\  479--535.
  Springer, 2012.

\bibitem[Melis et~al.(2017)Melis, Dyer, and Blunsom]{language_model_dm}
Melis, Gabor, Dyer, Chris, and Blunsom, Phil.
\newblock On the state of the art of evaluation in neural language models.
\newblock \emph{arXiv preprint arXiv:1707.05589}, 2017.

\bibitem[Mikolov et~al.(2010)Mikolov, Karafiat, Burget, Cernocky, and
  Sanjeev]{ptb_mikolov}
Mikolov, Thomas, Karafiat, Martin, Burget, Lukas, Cernocky, Jan, and Sanjeev,
  Khudanpur.
\newblock Recurrent neural network based language model.
\newblock In \emph{INTERSPEECH}, 2010.

\bibitem[Neelakantan et~al.(2016)Neelakantan, Vilnis, Le, Sutskever, Kaiser,
  Kurach, and Martens]{arvind_noise}
Neelakantan, Arvind, Vilnis, Luke, Le, Quoc, Sutskever, Ilya, Kaiser, Lukasz,
  Kurach, Karol, and Martens, James.
\newblock Adding gradient noise improves learning for very deep networks.
\newblock \emph{arXiv preprint arXiv:1511.06807}, 2016.

\bibitem[Orchard \& Wang(2016)Orchard and Wang]{generalized_neural_rule}
Orchard, Jeff and Wang, Lin.
\newblock The evolution of a generalized neural learning rule.
\newblock In \emph{2016 International Joint Conference on Neural Networks
  (IJCNN)}, pp.\  4688--4694, 2016.

\bibitem[Pascanu \& Bengio(2013)Pascanu and Bengio]{pascanu2013revisiting}
Pascanu, Razvan and Bengio, Yoshua.
\newblock Revisiting natural gradient for deep networks.
\newblock \emph{arXiv preprint arXiv:1301.3584}, 2013.

\bibitem[Pascanu et~al.(2013)Pascanu, Mikolov, and
  Bengio]{pascanu2013difficulty}
Pascanu, Razvan, Mikolov, Tomas, and Bengio, Yoshua.
\newblock On the difficulty of training recurrent neural networks.
\newblock In \emph{International Conference on Machine Learning}, 2013.

\bibitem[Press \& Wolf(2017)Press and Wolf]{weight_press}
Press, Ofir and Wolf, Lior.
\newblock Using the output embedding to improve language models.
\newblock \emph{arXiv preprint arXiv:1707.05589}, 2017.

\bibitem[Ravi \& Larochelle(2017)Ravi and Larochelle]{ravi2017opt}
Ravi, Sachin and Larochelle, Hugo.
\newblock Optimization as a model for few-shot learning.
\newblock In \emph{International Conference on Learning Representations}, 2017.

\bibitem[Riedmiller \& Braun(1992)Riedmiller and Braun]{riedmiller1992rprop}
Riedmiller, Martin and Braun, Heinrich.
\newblock {RPROP} - a fast adaptive learning algorithm.
\newblock In \emph{Proc. of ISCIS VII, Universitat}. Citeseer, 1992.

\bibitem[Runarsson \& Jonsson(2000)Runarsson and Jonsson]{evolve_rules}
Runarsson, Thomas~P. and Jonsson, Magnus~T.
\newblock Evolution and design of distributed learning rules.
\newblock In \emph{IEEE Symposium on Combinations of Evolutionary Computation
  and Neural Networks}, 2000.

\bibitem[Schaul et~al.(2013)Schaul, Zhang, and LeCun]{schaul2013no}
Schaul, Tom, Zhang, Sixin, and LeCun, Yann.
\newblock No more pesky learning rates.
\newblock In \emph{International Conference on Machine Learning}, 2013.

\bibitem[Schmidhuber(1992)]{schmid92}
Schmidhuber, Juergen.
\newblock Steps towards 'self-referential' neural learning: A thought
  experiment.
\newblock Technical report, University of Colorado Boulder, 1992.

\bibitem[Schraudolph(2002)]{schraudolph2002fast}
Schraudolph, Nicol~N.
\newblock Fast curvature matrix-vector products for second-order gradient
  descent.
\newblock \emph{Neural Computation}, 14\penalty0 (7):\penalty0 1723--1738,
  2002.

\bibitem[Schulman et~al.(2017)Schulman, Wolski, Dhariwal, Radford, and
  Klimov]{schulmanppo}
Schulman, John, Wolski, Filip, Dhariwal, Prafulla, Radford, Alec, and Klimov,
  Oleg.
\newblock Proximal policy optimization algorithms.
\newblock \emph{CoRR}, 2017.

\bibitem[Shazeer et~al.(2017)Shazeer, Mirhoseini, Maziarz, Davis, Le, Hinton,
  and Dean]{largenetworks}
Shazeer, Noam, Mirhoseini, Azalia, Maziarz, Krzysztof, Davis, Andy, Le, Quoc,
  Hinton, Geoffrey, and Dean, Jeff.
\newblock Outrageously large neural networks: The sparsely-gated
  mixture-of-experts layer.
\newblock In \emph{International Conference on Learning Representations}, 2017.

\bibitem[Wichrowska et~al.(2017)Wichrowska, Maheswaranathan, Hoffman,
  Colmenarejo, Denil, de~Freitas, and Sohl-Dickstein]{wichrowska2017}
Wichrowska, Olga, Maheswaranathan, Niru, Hoffman, Matthew~W., Colmenarejo,
  Sergio~Gomez, Denil, Misha, de~Freitas, Nando, and Sohl-Dickstein, Jascha.
\newblock Learned optimizers that scale and generalize.
\newblock In \emph{International Conference on Machine Learning}, 2017.

\bibitem[Williams(1992)]{Williams92simplestatistical}
Williams, Ronald~J.
\newblock Simple statistical gradient-following algorithms for connectionist
  reinforcement learning.
\newblock In \emph{Machine Learning}, 1992.

\bibitem[Wu et~al.(2016)Wu, Schuster, Chen, Le, Norouzi, Macherey, Krikun, Cao,
  Gao, Macherey, Klingner, Shah, Johnson, Liu, Kaiser, Gouws, Kato, Kudo,
  Kazawa, Stevens, Kurian, Patil, Wang, Young, Smith, Riesa, Rudnick, Vinyals,
  Corrado, Hughes, and Dean]{wu2016google}
Wu, Yonghui, Schuster, Mike, Chen, Zhifeng, Le, Quoc~V., Norouzi, Mohammad,
  Macherey, Wolfgang, Krikun, Maxim, Cao, Yuan, Gao, Qin, Macherey, Klaus,
  Klingner, Jeff, Shah, Apurva, Johnson, Melvin, Liu, Xiaobing, Kaiser, Lukasz,
  Gouws, Stephan, Kato, Yoshikiyo, Kudo, Taku, Kazawa, Hideto, Stevens, Keith,
  Kurian, George, Patil, Nishant, Wang, Wei, Young, Cliff, Smith, Jason, Riesa,
  Jason, Rudnick, Alex, Vinyals, Oriol, Corrado, Greg, Hughes, Macduff, and
  Dean, Jeffrey.
\newblock Google's neural machine translation system: Bridging the gap between
  human and machine translation.
\newblock \emph{arXiv preprint arXiv:1609.08144}, 2016.

\bibitem[Zagoruyko \& Komodakis(2016)Zagoruyko and
  Komodakis]{zagoruyko2016wide}
Zagoruyko, Sergey and Komodakis, Nikos.
\newblock Wide residual networks.
\newblock \emph{arXiv preprint arXiv:1605.07146}, 2016.

\bibitem[Zeiler(2012)]{zeiler2012adadelta}
Zeiler, Matthew~D.
\newblock Adadelta: an adaptive learning rate method.
\newblock \emph{arXiv preprint arXiv:1212.5701}, 2012.

\bibitem[Zhang et~al.(2017)Zhang, Bengio, Hardt, Recht, and
  Vinyals]{zhang2016understanding}
Zhang, Chiyuan, Bengio, Samy, Hardt, Moritz, Recht, Benjamin, and Vinyals,
  Oriol.
\newblock Understanding deep learning requires rethinking generalization.
\newblock In \emph{International Conference on Learning Representations}, 2017.

\bibitem[Zoph \& Le(2017)Zoph and Le]{ZophLe}
Zoph, Barret and Le, Quoc~V.
\newblock {Neural Architecture Search} with reinforcement learning.
\newblock In \emph{International Conference on Learning Representations}, 2017.

\bibitem[Zoph et~al.(2017)Zoph, Vasudevan, Shlens, and Le]{zoph2017learning}
Zoph, Barret, Vasudevan, Vijay, Shlens, Jonathon, and Le, Quoc~V.
\newblock Learning transferable architectures for scalable image recognition.
\newblock \emph{arXiv preprint arXiv:1707.07012}, 2017.

\end{thebibliography}
\bibliographystyle{icml2017}

\clearpage
\onecolumn

\section{\label{appendix} Appendix}

\subsection{Suggested usage for PowerSign and AddSign}

For the default hyperparameter values of the PowerSign and AddSign optimizers (i.e., $\alpha=e$ and $\alpha=1$ respectively), the learning rate used for SGD is usually a good first choice. 
As usual, when using learning rate decay, the initial learning rate should be larger than when not applying decay. When applying linear cosine learning rate decay, one can set up the initial learning rate to be larger than for cosine decay.
Concerning the internal decays which are applied to the $\sgn(g)*\sgn(m)$ quantity, we've generally found that PowerSign works best with internal cosine decay and AddSign with internal linear decay.

\subsection{Some additional update rules found by Neural Optimizer Search.}

\begin{table*}[h!]
\centering
\small
\begin{tabular}{l|cccc}
\toprule
\multicolumn{1}{c|}{\bf Optimizer} & {\bf Final Val} & {\bf Final Test} & {\bf Best Val} & {\bf Best Test}\\ 
\midrule
SGD & 92.0 & 91.8 & 92.9 & 91.9 \\
Momentum & 92.7 & 92.1 & 93.1 & 92.3 \\
Adam & 90.4 & 90.1 & 91.8 & 90.7 \\
RMSProp & 90.7 & 90.3 & 91.4 & 90.3 \\
\midrule
$\hat{m} * (1 + \epsilon) * sigmoid(10^{-4}w)$ & 90.6 & 90.6 & 93.1 & 92.2 \\
$\sgn(m)*\sqrt{|g|}$ & 92.2 & 91.8 & 92.9 & 92.2 \\
$\sgn(g)*\sgn(m)*\hat{m}$ & 91.2 & 91.0 & 92.4 & 91.3 \\ 
$\sgn(m)*\sqrt{|g|}$ & 91.7 & 91.1 & 92.3 & 91.6 \\
$(1 + \sgn(g)*\sgn(m))*\sgn(g)$ & 91.3 & 90.4 & 91.9 & 91.1 \\
$(1 + \sgn(g)*\sgn(m))*\sgn(m)$ & 91.0 & 90.6 & 92.0 & 90.8 \\
$\sgn(g)*\sqrt{|\hat{m}|}$ & 90.7 & 90.6 & 91.5 & 90.6 \\
$\sqrt{|g|}*\hat{m}$ & 92.0 & 90.9 & 93.6 & 93.1 \\
$\sqrt{|g|}*g$ & 92.6 & 91.9 & 93.2 & 92.3 \\
$(1 + \sgn(g)*\sgn(m))*\hat{m}$ & 91.8 & 91.3 & 92.6 & 91.8 \\
$(1 + \sgn(g)*\sgn(m))*\rmsprop$ & 92.0 & 92.1 & 92.9 & 92.4 \\
$(1 + \sgn(g)*\sgn(m))*\Adam$ & 91.2 & 91.2 & 92.2 & 91.9 \\
$[e^{\sgn(g)*\sgn(m)} + \clip(g, 10^{-4})]*g$ & 92.5 & 92.4 & 93.8 & 93.1 \\
$\clip(\hat{m}, 10^{-4}) * e^{\hat{v}}$ & 93.5 & 92.5 & 93.8 & 92.7 \\
$\hat{m} * e^{\hat{v}}$ & 93.1 & 92.4 & 93.8 & 92.6 \\
$g * e^{\sgn(g)*\sgn(m)}$ & 93.1 & 92.8 & 93.8 & 92.8 \\
$\drop(g, 0.3) * e^{\sgn(g)*\sgn(m)}$ & 92.7 & 92.2 & 93.6 & 92.7 \\
$\hat{m} * e^{g^2}$ & 93.1 & 92.5 & 93.6 & 92.4 \\
$\drop(\hat{m}, 0.1)/(e^{g^2}+\epsilon)$ & 92.6 & 92.4 & 93.5 & 93.0 \\
$\drop(g, 0.1) * e^{\sgn(g)*\sgn(m)}$ & 92.8 & 92.4 & 93.5 & 92.2 \\
$\clip(\rmsprop, 10^{-5}) + \drop(\hat{m}, 0.3)$ & 90.8 & 90.8 & 91.4 & 90.9 \\
$\Adam * e^{\sgn(g)*\sgn(m)}$ & 92.6 & 92.0 & 93.4 & 92.0 \\
$\Adam * e^{\hat{m}}$ & 92.9 & 92.8 & 93.3 & 92.7 \\
$g + \drop(\hat{m}, 0.3)$ & 93.4 & 92.9 & 93.7 & 92.9 \\
$\drop(\hat{m}, 0.1)*e^{g^3}$ & 92.8 & 92.7 & 93.7 & 92.8 \\
$g - \clip(g^2, 10^{-4})$ & 93.4 & 92.8 & 93.7 & 92.8 \\
$e^g-e^{\hat{m}}$ & 93.2 & 92.5 & 93.5 & 93.1 \\
$\drop(\hat{m}, 0.3) * e^{10^{-3} w}$ & 93.2 & 93.0 & 93.5 & 93.2 \\
\bottomrule
\end{tabular}

\caption{Performance of Neural Optimizer Search and standard optimizers on the Wide-ResNet architecture~\cite{zagoruyko2016wide} on CIFAR-10. Final Val and Final Test refer to the final validation and test accuracy after for training for 300 epochs. Best Val corresponds to the best validation accuracy over the 300 epochs and Best Test is the test accuracy at the epoch where the validation accuracy was the highest. For each optimizer we report the best results out of seven learning rates on a logarithmic scale according to the validation accuracy. No learning rate decay is applied.}
\label{tab:cifar10_appendix}
\end{table*}

\end{document}